%% file: egpaper_for_review.tex
\documentclass[10pt,twocolumn,letterpaper]{article}

\usepackage{iccv}
\usepackage{times}
\usepackage{epsfig}
\usepackage{graphicx}
\usepackage{amsmath}
\usepackage{amssymb}

\usepackage{float}
\usepackage[utf8]{inputenc}
\usepackage{makeidx}
\usepackage{url}
\usepackage{epstopdf}
\usepackage{doc}
\usepackage{enumerate}
\usepackage{bbm}
\usepackage{multicol, blindtext}
\usepackage{ctable}
\usepackage{textcomp}			
\usepackage[singlelinecheck=false]{caption}
\usepackage[]{footmisc}
\usepackage{multirow}
\usepackage{subfigure}
\usepackage{lipsum}

\newcommand\Tstrut{\rule{0pt}{2.6ex}}         
\newcommand\Bstrut{\rule[-0.9ex]{0pt}{0pt}}   

\usepackage{algorithm}
\usepackage{algpseudocode}
\usepackage{amsmath}

\iccvfinalcopy 

\def\iccvPaperID{5} 

\ificcvfinal\fi
\begin{document}

\title{Talking With Your Hands: Scaling Hand Gestures and Recognition With CNNs}

\author{\parbox{16cm}{\centering
    {\large Okan K\"op\"ukl\"u$^1$, Yao Rong$^{1,2}$, Gerhard Rigoll$^1$}\\
    {\normalsize
    \vspace{0.2cm}
    $^1$ Institute for Human-Machine Communication, TU Munich, Germany\\
    $^2$ Infineon Technologies AG, Germany}}
}

\maketitle
\thispagestyle{empty}

\input{abstract.tex}
\input{introduction.tex}

\input{related_work.tex}
\input{methodology.tex}
\input{experiments.tex}

\input{conclusion.tex}
\input{acknowledgements.tex}

{\small
\bibliographystyle{ieee}
\bibliography{egbib}
}


\end{document}

%% file: abstract.tex
\begin{abstract}

The use of hand gestures provides a natural alternative to cumbersome interface devices for Human-Computer Interaction (HCI) systems. As the technology advances and communication between humans and machines becomes more complex, HCI systems should also be scaled accordingly in order to accommodate the introduced complexities. In this paper, we propose a methodology to scale hand gestures by forming them with predefined gesture-phonemes, and a convolutional neural network (CNN) based framework to recognize hand gestures by learning only their constituents of gesture-phonemes. The total number of possible hand gestures can be increased exponentially by increasing the number of used gesture-phonemes. For this objective, we introduce a new benchmark dataset named Scaled Hand Gestures Dataset (SHGD) with only gesture-phonemes in its training set and 3-tuples gestures in the test set. In our experimental analysis, we achieve to recognize hand gestures containing one and three gesture-phonemes with an accuracy of 98.47\% (in 15 classes) and 94.69\% (in 810 classes), respectively. Our dataset, code and pretrained models are publicly available \footnote{https://www.mmk.ei.tum.de/shgd/}.

\end{abstract}

%% file: introduction.tex
\section{Introduction}

\begin{figure}[t!]
	\centering
	\includegraphics[width = 0.45\textwidth]{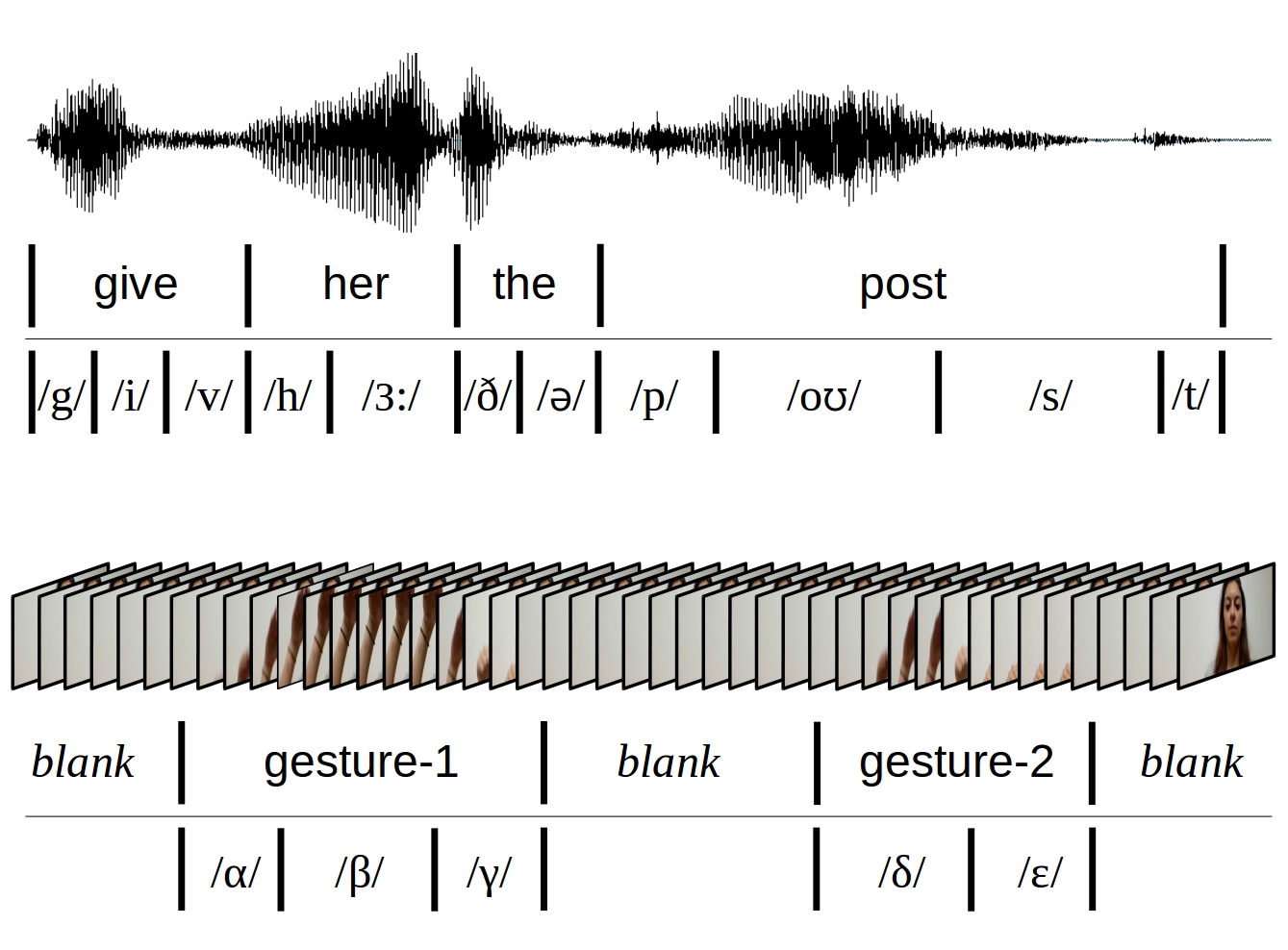}
	\caption{Top: An audio signal corresponding to the sentence ``give her the post". Each word in this sentence consists of one or multiple phonemes. Bottom: A video signal (i.e. sequence of frames) containing 2 hand gestures. Similar to speech signal, each gesture consists of one or multiple gesture-phonemes denoted by $\alpha, \beta, \gamma, \delta, \varepsilon$. The signals and their annotations are for illustrative purposes only.}
	\label{fig:speech_vs_gesture}
\end{figure}

Computers have become an indispensable part of human life. Therefore, facilitating natural human-computer interaction (HCI) contains utmost importance to bridge human-computer barrier. Gestures have long been considered as an interaction technique delivering natural and intuitive experience while communicating with computers. This is a driving force in the research community to work on gesture representations, recognition techniques and frameworks.  

As technology keeps advancing, the use of computers in our lives increases as well with additional new devices such as smart phones, watches, TVs, headphones, autonomous cars etc. Therefore, the communication between humans and machines gradually becomes more complex, requiring HCI systems to accommodate the introduced complexities.

In this work, we propose an approach to scale hand gestures by composing each gesture with multiple gesture-phonemes. The main inspiration comes from the phonology and morphology of the spoken languages. Fig. \ref{fig:speech_vs_gesture} (top) shows the morphological and phonological analysis of the sentence ``give her the post". Each word in this sentence is composed of a sequence of phonemes. Similarly, we create hand gestures using one or multiple gesture-phonemes sequentially, as shown in Fig. \ref{fig:speech_vs_gesture} (bottom). So, our motivation is first to learn the gesture-phonemes successfully, then to recognize hand gestures, which contains multiple gesture-phonemes, with only this knowledge. 

Structuring hand gestures with this approach enables to scale hand gestures without requiring to collect additional training data. For a given number of gesture-phonemes, the number of all possible hand gestures is exponentially proportional to the number of gesture-phonemes each gesture contains. For the proposed gesture scaling approach, we present a convolutional neural network (CNN) based framework using sliding-window approach together with Viterbi-like \cite{viterbi1967error} decoder algorithm. For the CNN model, we have used 2-dimensional (2D) and 3-dimensional (3D) SqueezeNet and MobileNetV2 models. 

This paper presents the following contributions:

\begin{enumerate}[(i)]
\setlength{\itemsep}{1pt}
\setlength{\parskip}{1pt}
    \item Our major contribution is creating hand gesture recognition framework, which is ``scalable" according to the complexity of the desired HCI system. To the best of our knowledge, this is the first work that address the scalability of hand gestures. The CNN model is only trained with 10 gesture phonemes and 3 signaling classes (\textit{preparation}, \textit{retraction} and \textit{no-gesture}), and the framework can recognize scaled gesture tuples with 3 gesture phonemes (as in this paper) or more. Assumed that a HCI system with the recognition capability of 810 different gestures needs to be implemented. With the old fashioned way, you need to define 810 different hand gestures, collect enough training samples (400 training samples for each class), train an architecture to get desired accuracy (remember that for ChaLearn IsoGD \cite{wan2016chalearn}, the state-of-the-art accuracy is around 80\% for 249 classes). With this framework, you just need to train with 10 gesture phonemes and 3 signaling classes, then for 810 classes (3-tuple gestures) you can achieve around 95\% accuracy. Consider the situation when you need 65610 different gestures (5-tuple gestures). Approximately 25 million training samples are needed for the old fashioned way.
    
    \item The second contribution is the benchmark dataset named Scaled Hand Gestures Dataset (SHGD), which will be made publicly available. The videos are collected using a Time-of-Flight (ToF) based 3D Image Sensor, which is shown in Fig. \ref{fig:data_collection}. The dataset contains only gesture-phonemes in its training set. For the test set, SHGD contains gesture-phonemes and 3-tuple gestures.
    
    \item The third contribution of the paper is that with the designed Viterbi-like decoder, the performed 3-tuple gestures are recognized only once. This contains utmost importance for online HCI systems. Moreover, designed Viterbi-like decoder is very lightweight as HCI systems should be designed considering the memory and power budget of the HCI system. 
\end{enumerate}




%% file: related_work.tex
\section{Related Work}
\label{ch:Related Work}

Ever since AlexNet \cite{krizhevsky2012imagenet}, deep CNNs have dominated nearly all computer vision tasks. At first, CNNs have infiltrated to the image-based tasks due to the availability of only large scale image datasets such as ImageNet \cite{deng2009imagenet}. Afterwards, CNNs are also applied for video analysis tasks. However, as the first video datasets were comparatively small such as UCF-101 \cite{soomro2012ucf101} and HMDB \cite{kuehne2011hmdb}, all initial video analysis architectures are based on 2D CNNs which utilize transfer learning from ImageNet, such as \cite{simonyan2014two, karpathy2014large, wang2016temporal, donahue2015long}. With the availability of large-scale video datasets like Sports-1M \cite{karpathy2014large}, Kinetics \cite{carreira2017quo}, Jester \cite{jester}, this problem was solved and successful 3D CNNs could be trained from scratch without overfitting \cite{hara2018can}.

Since gestures provide a natural, creative and intuitive interaction experience for communication with computers, hand gesture recognition is one of the most popular video analysis tasks. Although there have been many approaches using hand-crafted features like orientation of histograms \cite{freeman1995orientation}, histogram of oriented gradients (HOG) \cite{prasuhn2014hog} or bag-of-features \cite{dardas2011real}, the state of the art hand gesture recognition architectures are based on CNNs \cite{kopuklu2018motion, molchanov2015multi, Molchanov_2015_CVPR_Workshops, molchanov2016online, kopuklu2019real}, similar to other computer vision tasks.

Until recently, the primary trend has been to make CNNs deeper and more complicated \cite{hu2018squeeze, he2016deep} in order to achieve higher classification performance. But the pursue of lightweight networks with high accuracy is now growing, as in many real-time applications like autonomous driving and robotics, where the computation capability of the platform is always limited. Therefore, there has been several resource efficient CNN architectures such as SqueezeNet\cite{iandola2016squeezenet}, MobileNet \cite{howard2017mobilenets}, MobileNetV2 \cite{sandler2018mobilenetv2}, ShuffleNet \cite{Zhang2018ShuffleNetAE} and ShuffleNetV2 \cite{ma2018shufflenet}, which aim to reduce computational cost but still keep the accuracy high. In our work, we have used the 2D and 3D versions of SqueezeNet and MobileNetV2 since we want a lightweight framework.

Fusion of different modalities is another strategy that helps CNNs to improve recognition performance. However, fusion also introduces extra computational cost especially at decision \cite{simonyan2014two} and feature \cite{miao2017multimodal} level. On the other hand, \cite{kopuklu2018motion} proposes a data level fusion strategy, Motion Fused Frames (MFFs), where different modalities can be fused with very little modification to the network and computational cost. Since we have infrared (IR) and depth modalities in our dataset, we have adapted data level fusion strategy.

Although there have been many gesture recognition approaches, the idea of scaling hand gestures is very new but also very important in order to create complex HCI systems. To the best of our knowledge, this is the first work that scales hand gestures. More importantly, besides scaling, we achieve very similar recognition performance for gesture-tuples (94.69\% accuracy for 810 classes) compared to single gestures (98.47\% accuracy for 15 classes).

%% file: methodology.tex
\section{Methodology}

In this section, we fist describe the collected dataset. Afterwards, we explain the details of the experimented framework with its 2D and 3D CNN architectures and Viterbi-like decoder. Finally, we give the training details.

\begin{figure}[t!]
	\centering
	\includegraphics[width = 0.48\textwidth]{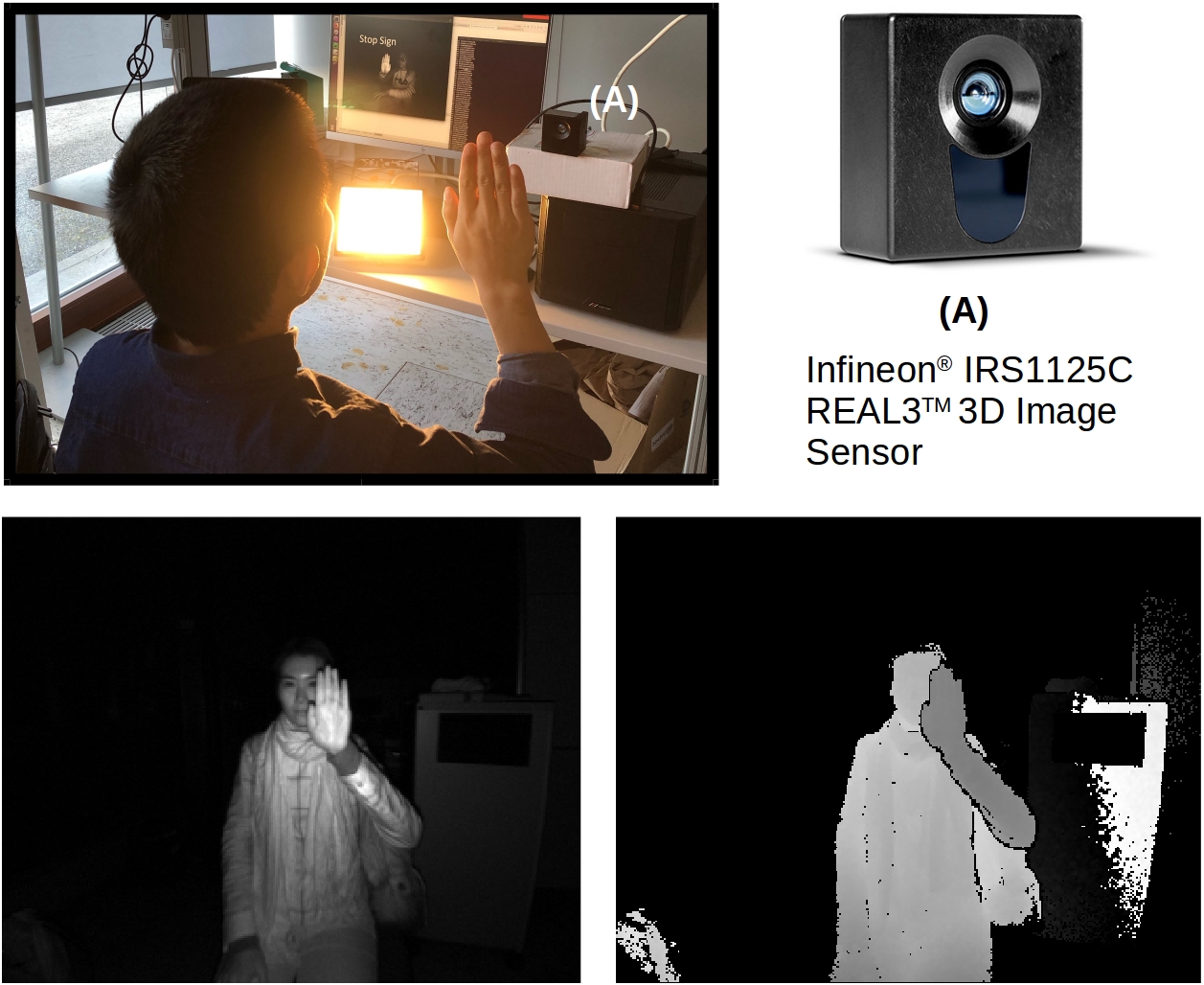}
	\caption{Data collection setup. Dataset is collected in infrared (bottom-left) and depth (bottom-right) modalities using Infineon\textsuperscript{\textregistered} IRS1125C REAL3\textsuperscript{TM} 3D Image Sensor.}
	\label{fig:data_collection}
\end{figure}

\subsection{Scaled Hand Gestures Dataset (SHGD)}

SHGD contains 15 single hand gestures, each recorded for infrared (IR) and depth modalities using Infineon\textsuperscript{\textregistered} IRS1125C REAL3\textsuperscript{TM} 3D Image Sensor. Each recording contains 15 gesture samples (one sample per class). There are in total 324 recordings from 27 distinct subjects in the dataset. Recordings of 8 subjects are reserved for testing, which makes 30\% of the dataset. Every subject makes 12 video recordings using two hands under 6 different environments, which are designed for increasing the network robustness against different lightning conditions and background disturbances. These environments are (1) indoors under normal daylight, (2) indoors under daylight and with an extra person in the background, (3) indoors at night under artificial lighting, (4) indoors in total darkness, (5) outdoors under intense sunlight and (6) outdoors under normal sunlight. We have simulated outdoor environments using two bright lights: Two lights for ``intense sunlight" and one light for ``normal sunlight". 

Fig. \ref{fig:data_collection} shows data collection setup, used camera and data samples. Subjects performed gestures while observing the computer screen, where the gestures were prompted in a random order. Videos are recorded at 45 frames per second (fps) with spatial resolution of 352$\times$287 pixels. Each recording lasts around 33 seconds.

\subsubsection{Single Gestures}

In its training set, SHGD contains only single gestures under 15 classes, which are given in Table \ref{gesture_table}. Recordings in the dataset are continuous video streams meaning that each recording contains \textit{no-gesture} and \textit{gesture} parts. Moreover, each \textit{gesture} contains \textit{preparation}, \textit{nucleus} and \textit{retraction} phases \cite{nespoulous2014biological, gavrila1999visual, gupta2016online}, which are critical for real-time gesture recognition.

\begin{table}[t!]
    \resizebox{\linewidth}{!}{
    \begin{tabular}{|c|l|c|l|c|l|}
    \hline
    \textbf{Label} & \textbf{Gesture} & \textbf{Label} & \textbf{Gesture}     & \textbf{Label} & \textbf{Gesture} \Tstrut\Bstrut  \\ \hline
    1     & Fist        & 6     & Two Fingers       & 11    & Swipe Left$^*$     \Tstrut\Bstrut \\\hline
    2     & Flat Hand   & 7     & Five Fingers      & 12    & Swipe Right$^*$    \Tstrut\Bstrut \\\hline
    3     & Thumb Up    & 8     & Stop Sign         & 13    & Pull Hand In$^*$   \Tstrut\Bstrut \\\hline
    4     & Thumb Left  & 9     & Check             & 14    & Move Hand Up$^*$   \Tstrut\Bstrut \\\hline
    5     & Thumb Right & 10    & Zero              & 15    & Move Hand Down$^*$ \Tstrut\Bstrut \\\hline
    \end{tabular}}
    \caption{15 single gesture classes in Scaled Hand Gesture Dataset (SHGD). \textbf{$^*$} marks the dynamic gestures which are not included as gesture-phonemes.}
    \label{gesture_table}
\end{table}

Among the single gesture classes listed in Table \ref{gesture_table}, static gestures are selected as gesture-phonemes since it is more convenient to perform different static gestures sequentially. For the rest of the paper, we will use the term \textit{phoneme} instead of \textit{gesture-phoneme} for the sake of easiness.

\subsubsection{Gesture Tuples}

Gesture tuple refers to hand gestures which contain sequentially performed phonemes. There are in total 10 different phonemes. When constructing gesture tuples, we leave out the consecutive same phonemes to avoid sequence length confusion. Therefore, the total number of different tuples can be calculated by the following equation:

\begin{equation}
    N=m(m-1)^{(s-1)}
\end{equation}

\noindent where $m$ is the number different phonemes and $s$ is the number of phonemes that the gesture tuple contains. 

Besides the test set for single gestures, SHGD also has a test set for gesture tuples containing 3 phonemes. 5 subjects perform gesture tuples under 5 different lightning conditions (excluding the environment of (2)). There are in total $10 \times (10 - 1)^{(3 - 1)} = 810$ permutations meaning different classes for 3-tuple gestures. Recordings are not segmented for this case. Therefore, one recording contains \textit{no-gesture}, \textit{3-tuple gesture} and \textit{no-gesture} without exact location of \textit{3-tuple gesture}. 

Since gestures are performed at different speeds in the real-life scenarios, we have also collected 3-tuple gestures at three different speeds: Slow, medium and fast. The subjects should finish 3-tuple gestures within 300 frames (6.7 sec), 240 frames (5.3 sec) and 180 frames (4 sec) for slow, medium and fast speed, respectively.

\subsubsection{SHGD-15 and SHGD-13}

SHGD-15 refers to the standard dataset where all single gestures in Table \ref{gesture_table} are included. On the other hand, SHGD-13 is specifically designed for 3-tuple gesture recognition. Besides 10 phonemes, SHGD-13 also contains \textit{preparation} (raising hand), \textit{retraction} (lowering hand) and \textit{no-gesture} classes. As there is no indication when a gesture starts and ends in the video, we use \textit{preparation} and \textit{retraction} classes to detect Start-of-Gesture (SoG) and End-of-Gesture (EoG). We use \textit{no-gesture} class to reduce the number false alarms since most of the time, ``no gesture" is performed in real-time gesture recognition applications \cite{kopuklu2019real}.

SHGD-15 is a balanced dataset with 96 samples in each class. However, SHGD-13 is an imbalanced dataset, where \textit{preparation} and \textit{retraction} classes contain 10 times more samples than phonemes, whereas \textit{no-gesture} contains around 20 times more samples than phonemes. Therefore, training of SHGD-13 requires special attention.

\subsection{Network Architecture}

\begin{figure}[t!]
	\centering
	\includegraphics[width = 0.48\textwidth]{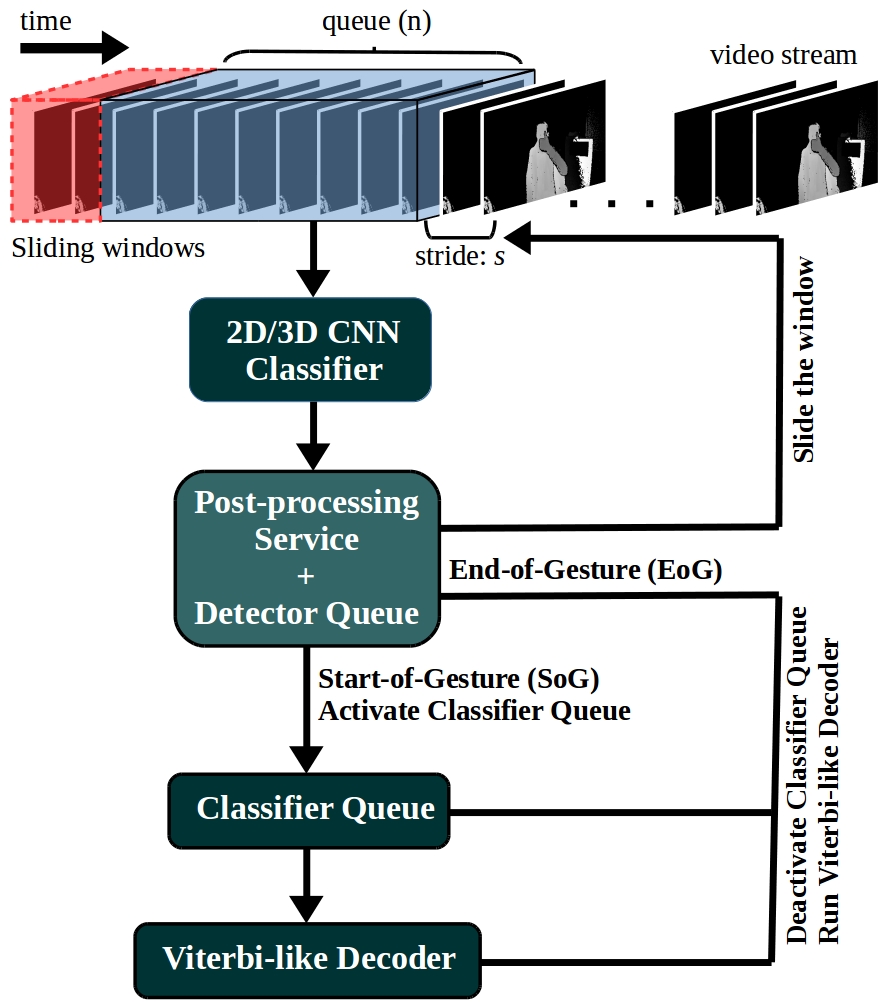}
	\caption{The general workflow of the proposed  architecture. Sliding windows with stride \textit{s} run through incoming video frames, and these frames in the queue are fed to a 2D or 3D CNN based classifier. The classifier's results are post-processed afterwards. After Start-of-Gesture (SoG) gets detected, the classifier queue is activated. Classifier's results are saved in the classifier queue until End-of-Gesture (EoG) is detected. Then, the Viterbi-like decoder runs on the classifier's queue to recognize the 3-tuple gesture.}
	\label{fig:arch}
\end{figure}

The general workflow of the proposed architecture is depicted in Fig. \ref{fig:arch}. A sliding window goes through the video stream with a queue size of 8 frames and stride \textit{s} of 1. The frames in the input queue is passed to a 2D/3D CNN which is pretrained on SHGD-13. The classification results are then post-processed by averaging with non-overlapping window size of 5. In this way, we can filter out some fluctuations due to the ambiguous states while changing the phonemes. Next, the post-processed outputs are fed into a detector queue, which tries to detect SoG and EoG. When the sum of class scores for \textit{preparation} is higher than the threshold, we set SoG flag on, activate the classifier queue and start storing the post-processed scores. Then, the detector queue is responsible for detecting EoG in a similar manner. After EoG flag is received, we deactivate the classifier queue and run the Viterbi-like decoder which recognizes the 3-tuple gesture. In the next parts, we explain the details for the main building blocks in the proposed architecture.

\subsubsection{2D and 3D CNN Classifiers}

CNN classifier is the most critical part of the proposed architecture. The properties of the deployed CNNs determine the detection and classification performance, memory usage and the speed of the overall architecture. In order to fulfill the resource constrained conditions and run as a real time application, two lightweight models are preferred selecting SqueezeNet \cite{iandola2016squeezenet} and MobileNetV2 \cite{sandler2018mobilenetv2} as classifiers in our architecture. In our analyses, we have deployed the 2D and 3D versions of these models.

The input to the CNN classifier is always 8 frames. Using these 8 frames, CNN classifier should recognize static phonemes together with dynamic preperation and retraction classes successfully. 3D CNNs can capture this dynamic motion information inherently due to their 3D convolutional kernels. However, 2D CNNs requires an extra spatiotemporal modeling in order to reason the relations between different frames.

\begin{figure}[b!]
	\centering
	\includegraphics[width=0.4\textwidth]{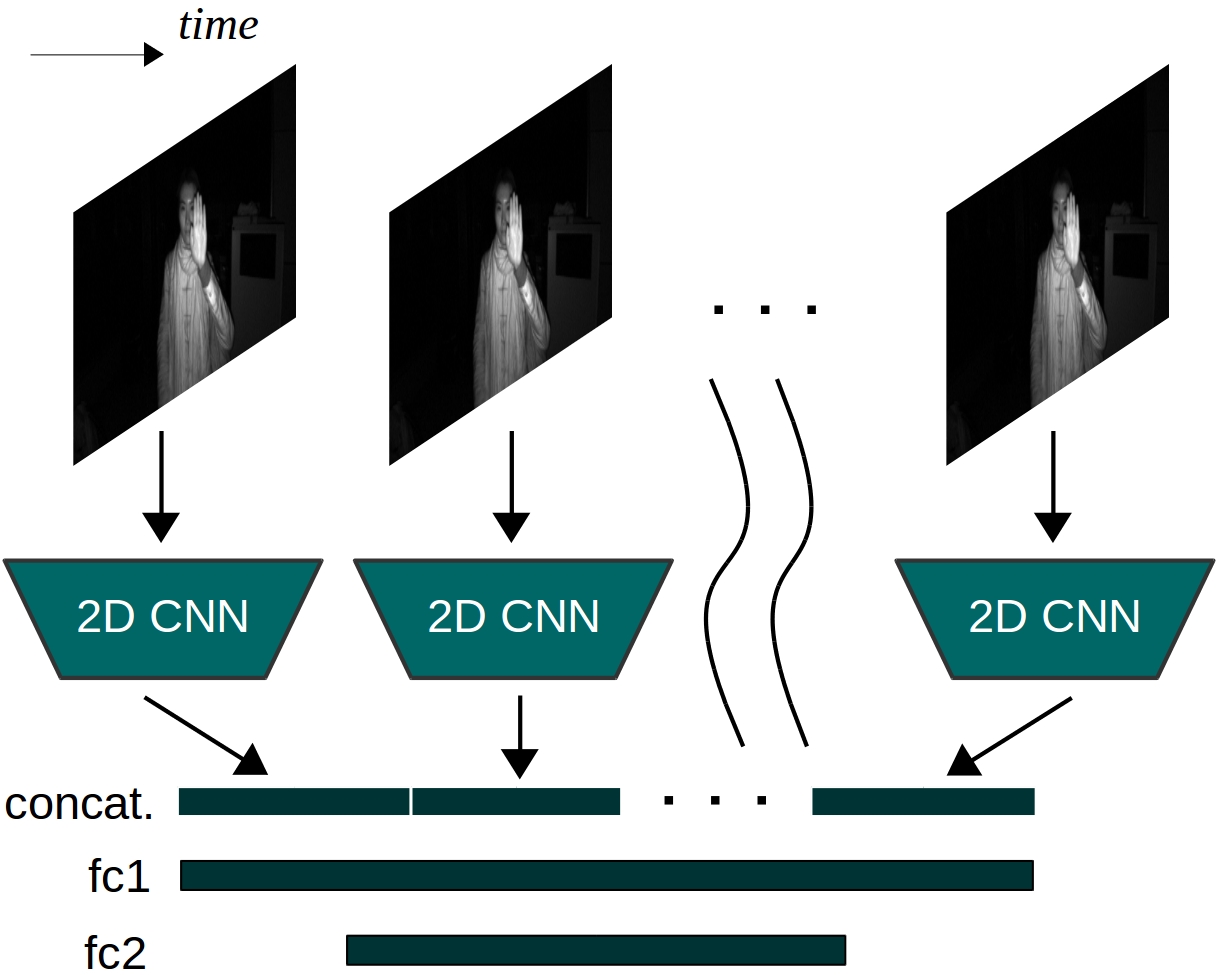}
	\caption{Spatiotemporal modeling approach used for 2D CNN models.}
	\label{fig:fusion}
\end{figure}

Fig. \ref{fig:fusion} depicts the applied spatiotemporal modeling approach used for 2D CNN models. Features of each 8 frames are extracted using the same 2D CNN and concatenated keeping their order intact. Afterwards, two levels of fully connected (fc) layers are applied in order to get class-conditional probability scores. The reason behind is that fc layers can organically infer the temporal relations, without knowing it is a sequence at all. The size of features 2D CNNs extracts is 64 for each frame. With the first fc layer, feature dimension is reduced from 64$\times$8=512 to 256. With the second fc layer, dimension is reduced to the number of classes.

\begin{figure}[t!]
	\centering
	\includegraphics[width=0.5\textwidth]{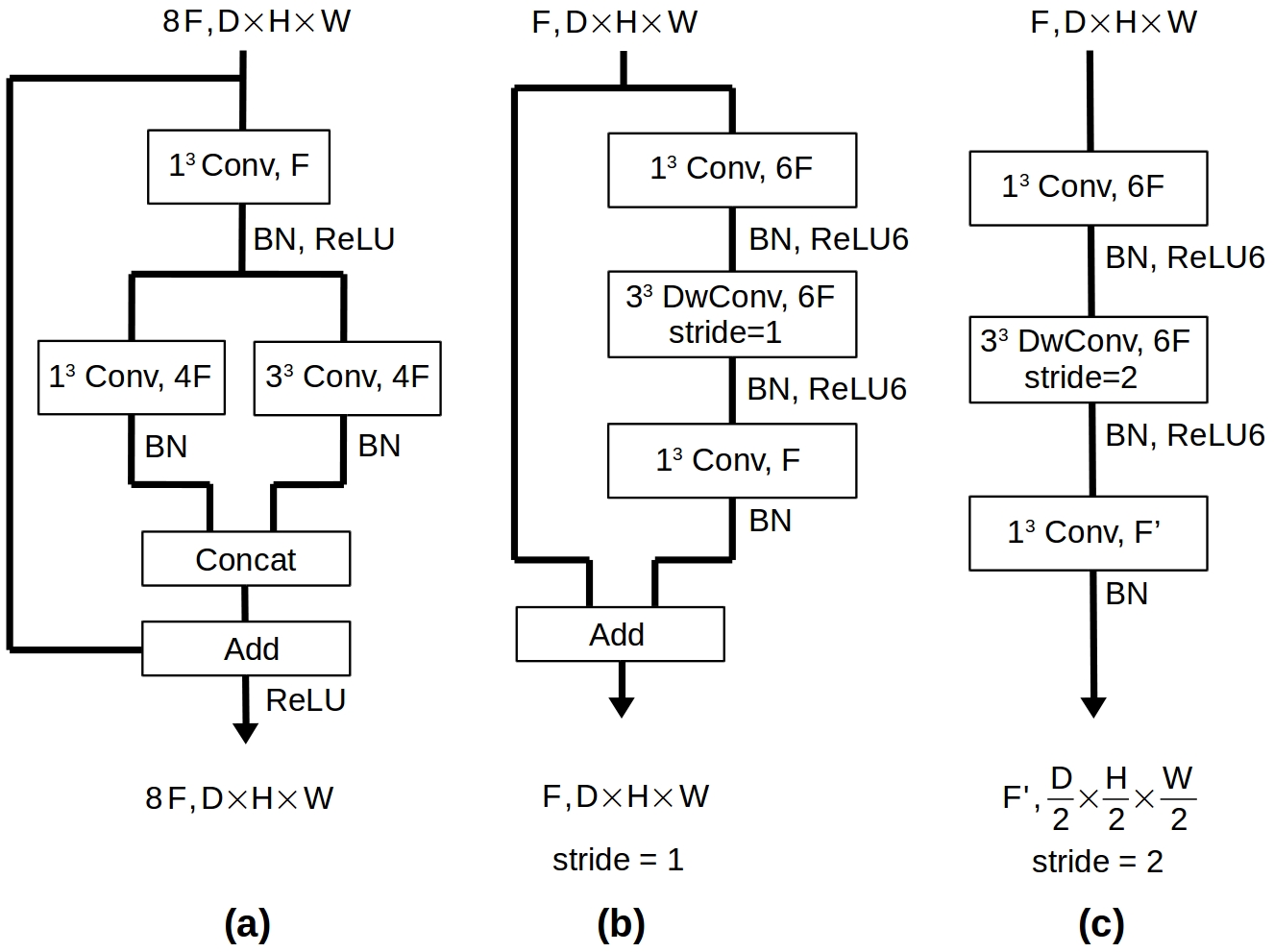}
	\caption{Blocks used in 3D CNN architectures. F is the number of feature maps and D$\times$H$\times$W stands for Depth$\times$Height$\times$Width for the input and output volumes. \mbox{DwConv} stands for depthwise convolution. 1$^{3}$ and 3$^{3}$ refers to kernel sizes of 1$\times$1$\times$1 and 3$\times$3$\times$3, respectively. \textbf{(a)} SqueezeNet's Fire block with simple bypass; \textbf{(b)} MobileNetV2's inverted residual block with stride 1; \textbf{(c)} MobileNetV2's inverted residual block with spatiotemporal downsampling (2$\times$).}
	\label{fig:blocks}
\end{figure}

On the other hand, 3D CNNs contains spatiotemporal modeling intrinsically and does not require an extra mechanism. We have inflated SqueezeNet and MobileNetV2 such that they accept 8 frames as input. The details of the 3D-SqueezeNet and 3D-MobileNetV2 are given in Table \ref{tab:3d squeezenet_arch} and Table \ref{tab:3d mobilenetv2_arch}, respectively. Their main building blocks are also depicted in Fig. \ref{fig:blocks}.

3D-SqueezeNet is deployed with simple bypass, as it achieves better results in the original architecture. However, we have not used simple bypass for its 2D version, as 2D-SqueezeNet pretrained on ImageNet is only available without bypass. For MobileNetV2, we have used $width\_multiplier$ of 1 for both 2D and 3D versions.

\begin{table}[t!]
	\centering
	\begin{tabular}{lll}
		\specialrule{.15em}{.0em}{.3em}
		\textbf{Layer / Stride}    & \textbf{Filter size}  & \textbf{Output size}\\ 
		\specialrule{.15em}{.3em}{.1em}
		Input clip            &         & $c$$\times$8$\times$112$\times$112  \\
		Conv1/s(1,2,2)        & 3$\times$3$\times$3   & 64$\times$8$\times$56$\times$56   \\
		MaxPool/s(1,2,2)      & 3$\times$3$\times$3   & 64$\times$8$\times$28$\times$28    \\
		\specialrule{.1em}{.1em}{.1em}
		Fire2                 &         & 128$\times$8$\times$28$\times$28   \\
		Fire3                 &         & 128$\times$8$\times$28$\times$28   \\
		MaxPool/s(2,2,2)      & 3$\times$3$\times$3   & 128$\times$4$\times$14$\times$14   \\
		\specialrule{.1em}{.1em}{.1em}
		Fire4                 &         & 256$\times$4$\times$14$\times$14   \\
		Fire5                 &         & 256$\times$4$\times$14$\times$14   \\
		MaxPool/s(2,2,2)      & 3$\times$3$\times$3   & 256$\times$2$\times$7$\times$7     \\
		\specialrule{.1em}{.1em}{.1em}
		Fire6                 &         & 384$\times$2$\times$7$\times$7     \\
		Fire7                 &         & 384$\times$2$\times$7$\times$7     \\
		MaxPool/s(2,2,2)      & 3$\times$3$\times$3   & 384$\times$1$\times$4$\times$4     \\
		\specialrule{.1em}{.1em}{.1em}
		Fire8                 &         & 512$\times$1$\times$4$\times$4     \\
		Fire9                 &         & 512$\times$1$\times$4$\times$4     \\
		\specialrule{.1em}{.1em}{.1em}
		Conv10/s(1,1,1)       & 1$\times$1$\times$1   & \textit{NumCls}$\times$1$\times$4$\times$4   \\
		AvgPool/s(1,1,1)      & 1$\times$4$\times$4   & \textit{NumCls} \\
		\specialrule{.15em}{.1em}{.0em}
	\end{tabular}
	\caption{3D-SqueezeNet architecture. Fire block is depicted in Fig. \ref{fig:blocks} (a).}
	\label{tab:3d squeezenet_arch}
\end{table}

\begin{table}[b!]
	\centering
	\begin{tabular}{lcl}
		\specialrule{.15em}{.0em}{.3em}
		\textbf{Layer / Stride}   & \textbf{Repeat}  & \textbf{Output size} \\ 
		\specialrule{.15em}{.3em}{.1em}
		Input clip              &           & $c$$\times$8$\times$112$\times$112   \\
		Conv1(3$\times$3$\times$3)/s(1,2,2)   & 1         & 32$\times$8$\times$56$\times$56    \\
		\specialrule{.1em}{.1em}{.1em}
	    Block/s(1,1,1)          & 1         & 16$\times$8$\times$56$\times$56    \\
		Block/s(1,2,2)          & 2         & 24$\times$8$\times$28$\times$28    \\
		Block/s(2,2,2)          & 3         & 32$\times$4$\times$14$\times$14    \\
		Block/s(2,2,2)          & 4         & 64$\times$2$\times$7$\times$7      \\
		Block/s(1,1,1)          & 3         & 96$\times$2$\times$7$\times$7      \\
		Block/s(2,2,2)          & 3         & 160$\times$1$\times$1$\times$1     \\
		Block/s(1,1,1)          & 1         & 320$\times$1$\times$1$\times$1     \\
		\specialrule{.1em}{.1em}{.1em}
		Conv(1$\times$1$\times$1)/s(1,1,1)    & 1         & 1280$\times$1$\times$1$\times$1    \\
		Linear(1280$\times$\textit{NumCls})    & 1    & \textit{NumCls} \\
		\specialrule{.15em}{.1em}{.0em}
	\end{tabular}
	\caption{3D-MobileNetV2 architecture. Block is inverted residual block whose details are given in Fig. \ref{fig:blocks} (b) and (c). Expansion factor of 6 is applied except for the initial Block where expansion factor of 1 is applied.}
	\label{tab:3d mobilenetv2_arch}
\end{table}

The spatial size of the inputs are 224 and 112 for 2D and 3D CNNs, respectively. The number of input channels $c$ depends on the experimented input data modality. Besides IR and depth, we have also applied data level fusion to IR and Depth (IR+D) in our experiments. We have used RGB modality only in pretrainings. Accordingly, the number of input channels are 3, 2, 1, 1 for RGB, IR+D, IR, depth modalities, respectively. The final size of inputs are $c$$\times$224$\times$224 for 2D CNNs, and $c$$\times$8$\times$112$\times$112 for 3D CNNs.

\subsubsection{Viterbi-like Decoder}

Viterbi decoding was invented by Andrew Viterbi \cite{viterbi1967error} and is now widely used in decoding convolutional codes. It is an elegant and efficient way to find out the optimal path with minimal error. In this paper, we have adapted it and used a Viterbi-like decoder to find out the phoneme sequences in 3-tuple gestures with maximal probability. Same as conventional Viterbi algorithm, we narrow down the optional paths systematically for each new input in the classifier queue.

For the Viterbi-like decoder, we introduced a couple of terms for better comprehensibility: \textit{K} is the number of allowed state transitions in the output sequence, which is 2 as we use 3-tuple gestures. The state refers to a phoneme in a path for the given time instant. \textit{P} refers to class-conditional probability scores for phonemes stored in Classifier Queue, which is shown in (2), whose columns $P_t$ are the average probability scores of each phoneme for five consecutive time instants. $P_t$ values are softmaxed before putting in \textit{P}. \textit{T} is the length of \textit{P} (i.e. number of columns), and \textit{N} is the number of phoneme classes, which is $10$ in our case. Therefore, the size of \textit{P} is \textit{T}$\times$\textit{N}.  
\begin{equation}
\label{Pmatrix}
P=\left[
\begin{matrix}
\bigm| & \cdots &\bigm| & \cdots & \bigm|\\
P_0 & \cdots &  P_t & \cdots  & P_{T-1} \\
\bigm| & \cdots  & \bigm| & \cdots & \bigm|\\
\end{matrix}
\right], \hspace{0.1cm}
P_{t} =
\left[
\begin{matrix}
p_{t,0} \\ p_{t,1}\\ \vdots \\ p_{t,N-1} 
\end{matrix}
\right]
\end{equation}
The probability of a path is the sum of the probability scores of all the states that this path goes through. Besides the number of allowed transitions \textit{K}, we introduce another constraint, transition cost $\delta$, in order to prevent false state transitions in the path. A path metric \textit{M} holds the paths $m_{t,i}$ with their sequence record $\pi_{t,i}$, path score $s_{t,i}$ and the transition times $k_{t,i}$. The path $m_{t,i}$ is shown as following:
\begin{equation}
    m_{t,i} = \left[ \pi_{t,i}, s_{t,i}, k_{t,i} \right], \hspace{0.3cm} 0\leq i < \gamma, \hspace{0.2cm}  0\leq t < T
\end{equation}
The state of path $m_{t,i}$ at time instant $t$ is denoted as $n_{t,i}$, and the last state in $\pi_{t,i}$ is also denoted as $\pi^{last}_{t,i}$. The transition cost is set to -0.2. The path scores \textit{s}, transition record $k$ and sequence record $\pi$ are updated with every new $P_t$ as following:
\thickmuskip=0mu
\begin{equation}
    s_{t+1,i}=s_{t,i}+p_{t+1,i}+\delta, \hspace{0.2cm}
    \delta=\begin{cases}
    -0.2, & \text{if} \hspace{0.05cm}  n_{t+1,i} \neq \pi^{last}_{t,i} \\
          & \text{and} \hspace{0.1cm} k_{t,i} < K \\
    0,    & \text{otherwise}         
  \end{cases}
\end{equation}

\begin{equation}
    \pi_{t+1,i}=\begin{cases}
  \pi_{t,i}\cup n_{t+1,i}, & \text{if} \hspace{0.1cm} n_{t+1,i} \neq \pi^{last}_{t,i} \hspace{0.05cm} \text{and} \hspace{0.05cm} k_{t,i}<K \\
  \pi_{t,i}, &\text{otherwise}
  \end{cases}
\end{equation}

\begin{equation}
     k_{t+1,i}=\begin{cases}
     k_{t,i}+1, & \text{if} \hspace{0.1cm} n_{t+1,i} \neq \pi^{last}_{t,i} \hspace{0.1cm} \text{and} \hspace{0.1cm} k_{t,i}<K \\
     k_{t,i}, &\text{otherwise}
  \end{cases}
\end{equation}

\begin{figure}[b!]
	\centering
	\includegraphics[width=0.5\textwidth]{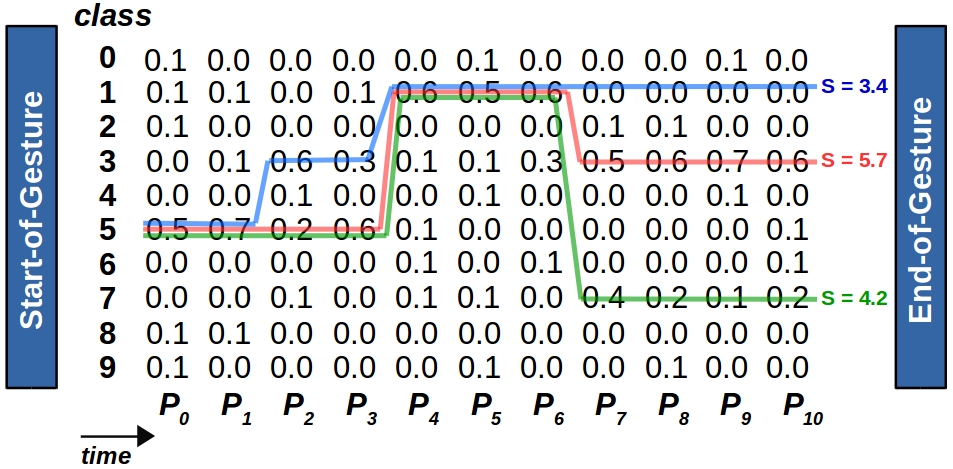}
	\caption{Illustration of our Viterbi-like decoder for 3-tuple gesture recognition. For the sake of simplicity, we have highlighted only three paths while the correct one is in red. For the correct path, $\pi$ = [5,1,3], s = 6.1 and k = 2. $2$ times the transition cost of $0.2$ is subtracted from each path.}
	\label{fig:vit_exp}
\end{figure}

In order to reduce computation, we limit the number of paths in \textit{M} to $\gamma$, which is set to 300. The working mechanism of the proposed Viterbi-like decoder is given in algorithm \ref{viterbi algorithm}. Fig. \ref{fig:vit_exp} depicts the illustration of our Viterbi-like decoder. Our decoder can inherently deal with the ambiguities at phoneme transitions as it naturally makes use of temporal ensembling.


\begin{algorithm}[t!]
  \caption{Viterbi-like decoder for 3-tuple gesture recognition}
  \label{viterbi algorithm}
  \begin{algorithmic}[1]
    \Function {Viterbi-like decoder}{$P,S$}
    \State Initialize  $s$, $\pi$ and $k$ at $P_0$;
    \For{each $P_t$}
    \State{Create all possible paths}
    \State{Update $s$, $\pi$ and $k$ according to (4), (5) and (6)}
    \State {Descending sort all $m$ in $M$ with their scores $s$}
    \State {Keep no more than the first $\gamma$ paths}
    \EndFor 
    \State \Return $\pi$ of $m$ with maximum $s$ and $k=K$
  \EndFunction
  \end{algorithmic}
\end{algorithm}

\subsection{Training Details}
\label{sec:training}

In the trainings, we have used Stochastic Gradient Descent (SGD) with standard categorical cross-entropy loss. While we have used 5$\times$10$^{-4}$ and 1$\times$10$^{-3}$ weight decay for 2D and 3D CNNs, respectively, the momentum is kept same as 0.9 for all the trainings. As Jester is the largest available hand gesture dataset \cite{jester}, we have pretrained all models on Jester dataset before fine tuning on \mbox{SHGD-15} and \mbox{SHGD-13}. For 2D CNN models, before Jester pretraining, we also have used models pretrained with ImageNet as starting point. The learning rate for 2D CNNs is initialized at 0.001 and reduced with a factor of 0.1 at 25$^{th}$, 35$^{th}$ and 45$^{th}$ epochs. For trainings of 3D CNNs on Jester dataset, learning rate is initialized with 0.1 and reduced twice with a factor of 0.1 at 30$^{th}$ and 45$^{th}$ epochs. All trainings are completed at $60^{th}$ epoch for Jester and SHGD.

For fine tuning of \mbox{SHGD-15} and \mbox{SHGD-13}, the pretrained parameters are loaded except for the first convolutional layer and the last fully connected layer. The number of input channels for the first convolutional layer is modified from 3 (RGB) to 2 for IR+D and 1 for IR and Depth modalities. In the last fully connected layer, the number of output features is set to the number of classes in SHGD. For \mbox{SHGD-13}, we have deployed weighted categorical cross-entropy loss as it is an unbalanced dataset.

We have deployed several data augmentation techniques such as random rotation ($\pm 10^\circ$), random resizing and random spatial cropping. Apart from spatial augmentations, we also applied temporal augmentations. Input clips are selected from random temporal positions given the bounds of each class. Moreover, at pretraining of 2D CNNs on Jester dataset, frames are selected randomly within each segment of videos as in Temporal Segment Network (TSN) \cite{wang2016temporal}, which introduces extra variation in the trainings.

%% file: experiments.tex
\section{Experiments}

\subsection{Results using Jester dataset}

Jester is currently the largest available hand gesture dataset. There are in total 148.092 video samples collected for 27 different classes. As the labels of the test set are not publicly available, we have experimented on the validation set of the dataset. Table \ref{tab:jester} summarizes the achieved results for our models. Besides the classification accuracy, the computational complexity in terms of floating point operations (FLOPs) and number of parameters are also given in Table \ref{tab:jester} in order to highlight the resource efficiency of our models. The best result is achieved by 3D-MobileNetV2 with accuracy of 93.33\%.

\begin{table}[t!]
	\centering
	\begin{tabular}{lccc}
		\specialrule{.15em}{.0em}{.1em}
		\textbf{Model}    & \textbf{Params}  & \textbf{MFLOPs} & \textbf{Acc.(\%)}\\ 
		\specialrule{.15em}{.1em}{.1em}
		2D-SqueezeNet       & 0.89M   & 310   & 87.40  \\
		2D-MobileNetV2      & 2.41M   & 366   & 91.35 \\
		3D-SqueezeNet       & 1.85M   & 686   & 87.74  \\ 
		3D-MobileNetV2      & 2.39M   & 344   & 93.33  \\
		\specialrule{.15em}{.1em}{.0em}
	\end{tabular}
	\caption{Results of different models on the validation set of Jester dataset. For 2D CNNs, FLOPs are calculated for extracting one frames features and final fc layers.}
	\label{tab:jester}
\end{table}

\subsection{Results using SHGD-15 and SHGD-13}

The performance of our models for SHGD-15 and SHGD-13 using different modalities are given in Table \ref{tab:SHGH}. The best results are achieved by 2D-SqueezeNet (98.47\%) and 3D-MobileNetV2 (96.06\%) for SHGD-15 and SHGD-13, respectively, both at IR+D modality.

\begin{figure}[h]
  \centering
\subfigure[]{
   \includegraphics[width=1.0\linewidth]{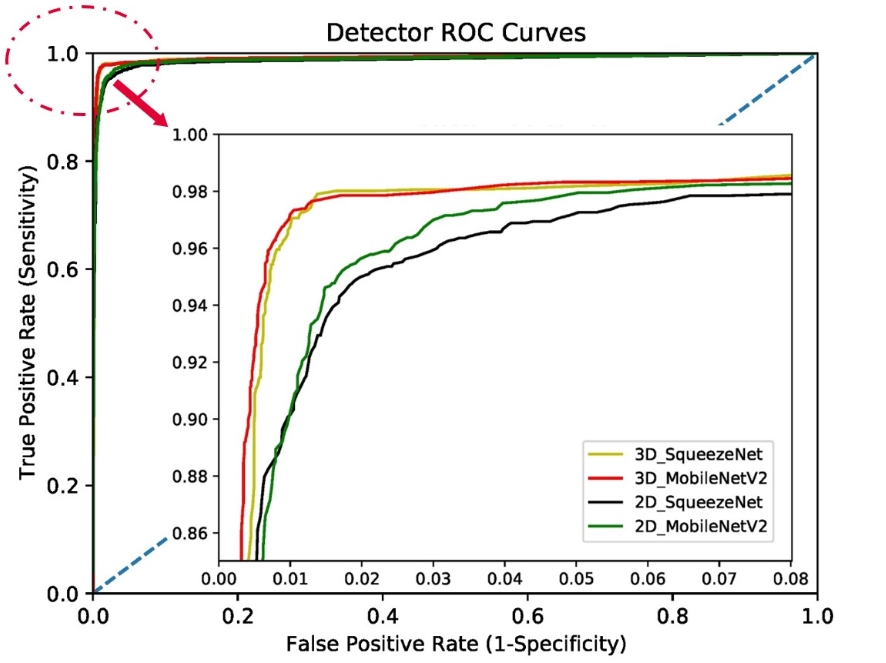}}
   \qquad
    \subfigure[]{
    \includegraphics[width=1.0\linewidth]{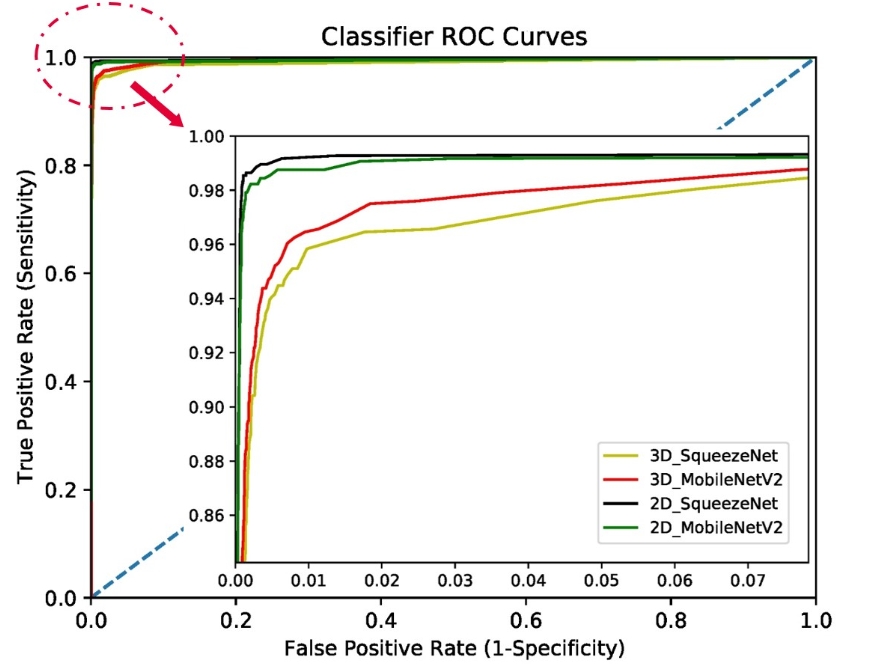}}
   \caption{ROC curves of 4 different models trained on SHGD-13 with IR+D modality. \textbf{(a)} Average ROC curves for dynamic preparation and retraction classes, \textbf{(b)} Average ROC curves of all the static phoneme classes.}
	\label{fig:roc_curves}
\end{figure}

For SHGD-15, 2D CNNs always achieve better results than 3D CNNs for all modalities. This is because of the fact that around 66.67\% of samples in SHGD-15 are static gestures, and 2D CNNs captures static content better than 3D CNNs. On the other hand, around 20\% of samples in SHGD-13 are static gestures resulting 3D CNNs to perform better. In order to highlight this situation, we have plotted the receiver operating characteristics (ROC) curves for static phoneme classes; and dynamic preparation and retraction classes in SHGD-13, which can be seen in Fig. \ref{fig:roc_curves}, where the same results can be observed.

Different models are sensitive to different data modalities. For instance, 2D-MobileNetV2 performs better at depth modality, whereas 3D-MobileNetV2 performs best at IR+D modality. However, fusion of different modalities (IR+D) results in better performance most of the time.

\begin{table}[t!]
    \centering
    \begin{tabular}{clccc}
    \specialrule{.15em}{.0em}{.1em}
        & \multirow{2}{*}{\textbf{Model}}  & \multicolumn{2}{c}{\textbf{Accuracy (\%)}} \\ \cmidrule(lr){3-4}
        &  & \textbf{SHGD-15} & \textbf{SHGD-13} \\     
        \specialrule{.15em}{.1em}{.1em}
        \multicolumn{1}{c|}{\multirow{4}{*}{\rotatebox[origin=c]{90}{\textbf{IR}}}} & 2D-SqueezeNet & 98.13 &  92.56\\  
        \multicolumn{1}{c|}{}  & 2D-MobileNetV2 & 97.36 & 93.11 \\ 
        \multicolumn{1}{c|}{}  & 3D-SqueezeNet  & 92.99& 95.87\\ 
        \multicolumn{1}{c|}{}  & 3D-MobileNetV2 & 92.85 & 94.62 \\  
        \specialrule{.15em}{.1em}{.1em}
        \multicolumn{1}{c|}{\multirow{4}{*}{\rotatebox[origin=c]{90}{\textbf{Depth}}}} & 2D-SqueezeNet & 98.13    & 95.02 \\  
        \multicolumn{1}{c|}{}  & 2D-MobileNetV2 & 98.13  & 95.64  \\ 
        \multicolumn{1}{c|}{}  & 3D-SqueezeNet  & 89.93 & 95.87\\ 
        \multicolumn{1}{c|}{}  & 3D-MobileNetV2 & 92.78  & 95.85\\  
        \specialrule{.15em}{.1em}{.1em}
        \multicolumn{1}{c|}{\multirow{4}{*}{\rotatebox[origin=c]{90}{\textbf{IR+D}}}} & 2D-SqueezeNet  & 98.47 & 93.94  \\
        \multicolumn{1}{c|}{}  & 2D-MobileNetV2 &  97.92 & 95.06 \\ 
        \multicolumn{1}{c|}{}  & 3D-SqueezeNet & 92.64 & 95.59  \\
        \multicolumn{1}{c|}{}  & 3D-MobileNetV2 & 94.31 & 96.06  \\ 
        \specialrule{.15em}{.1em}{.0em}
    \end{tabular}
    \caption{Results of different models with different modalities on the test sets of SHGD-15 and SHGD-13.}
	\label{tab:SHGH}
\end{table}

\subsection{Results for 3-tuple gesture recognition}

In this section, we evaluate the performance of our models for 3-tuple gesture recognition. Test set for this objective contains 1620 samples from 810 different permutations (i.e. classes). In order to evaluate the performance, three different errors and the total accuracy are defined as following: 

\begin{itemize}
    \item Detector error: The number of the gesture tuples, in which SoG or EoG is not successfully detected. It includes the flags detected at the wrong time and flags not detected at all. 
    \item Tuple error: The number of the gesture tuples, whose predicted sequence does not match to the ground truth.
    \item Single error: The number of the single phonemes which are recognized mistakenly inside the tuple error. For instance, if the ground truth is [6,8,10] and the recognized tuple is [6,10,12], then the single error is 2. 
    \item Total accuracy: The percentage of the correctly predicted tuples in the whole test set, where $N_{samples}$ is equal to 1620. It is calculated as following:
    \begin{equation}
        Acc = (1-\frac{Err_{det} + Err_{tup}}{N_{samples}}) \%
    \end{equation}
\end{itemize}

For this task, models are trained with SHGD-13. Table \ref{tab:tuple} gives the performance of experimented models on different modalities for 3-tuple gesture recognition. For the detection threshold of detector, 5 and 6 are used for 2D and 3D CNNs, respectively. Similar to previous results, 3D CNNs capture dynamic classes better and make less detector errors. On the other hands, 2D CNNs make less tuple and single error as they consist of static classes.

3D-MobileNetV2 achieves the best performance with an accuracy of 94.69\% for recognizing 810 different gesture tuples. 3D CNNs surpass 2D CNNs in this task generally, except for depth modality. We assume that this is due to the noise pixels appearing in depth modality from time to time. Therefore, 3D CNNs fail to capture the temporal relations between noisy frames.

\begin{table}[t!]
    \centering
    \begin{tabular}{clcccc}
    \specialrule{.15em}{.0em}{.1em}
        & \multirow{2}{*}{\textbf{Model}}  & \multicolumn{3}{c}{\textbf{Error}} & \multirow{2}{*}{\textbf{Acc.(\%)}} \\ \cmidrule(lr){3-5}
        &   & \textbf{Det} & \textbf{Tup} & \textbf{Sin} \\     
        \specialrule{.15em}{.1em}{.1em}
        \multicolumn{1}{c|}{\multirow{4}{*}{\rotatebox[origin=c]{90}{\textbf{IR}}}} & 2D-SqueezeNet & 191 & 54 &126 & 84.88\\
        \multicolumn{1}{c|}{}& 2D-MobileNetV2& 116 & 103 & 248 & 86.60    \\ 
        \multicolumn{1}{c|}{}& 3D-SqueezeNet & 11 & 159 & 375 & 89.51   \\ 
        \multicolumn{1}{c|}{}& 3D-MobileNetV2 & 10 & 209 & 492 & 86.48 \\
        \specialrule{.15em}{.1em}{.1em}
        \multicolumn{1}{c|}{\multirow{4}{*}{\rotatebox[origin=c]{90}{\textbf{Depth}}}}&2D-SqueezeNet & 73 & 127 & 275& 87.65\\
        \multicolumn{1}{c|}{}& 2D-MobileNetV2 & 77 & 111 & 259 & 88.40      \\ 
        \multicolumn{1}{c|}{} & 3D-SqueezeNet & 68 & 200 & 261 & 83.46
        \\ 
        \multicolumn{1}{c|}{} & 3D-MobileNetV2& 82 & 169 & 271 & 84.51  \\
        \specialrule{.15em}{.1em}{.1em}
        \multicolumn{1}{c|}{\multirow{4}{*}{\rotatebox[origin=c]{90}{\textbf{IR+D}}}} & 2D-SqueezeNet & 125 & 79 & 184 & 87.41 \\  
        \multicolumn{1}{c|}{} & 2D-MobileNetV2 & 41 & 71 & 165 & 93.09\\ 
        \multicolumn{1}{c|}{} & 3D-SqueezeNet  & 7 & 103 &  228  & 93.21 \\ 
        \multicolumn{1}{c|}{} & 3D-MobileNetV2 & 3  & 83  & 171 & 94.69  \\
        \specialrule{.15em}{.1em}{.0em}
    \end{tabular}
    \caption{Performance for the tuple detection. Det, Tup and Sin refer to the number of detector, tuple and single phoneme errors out of 1620 test samples.}
	\label{tab:tuple}
\end{table}

%% file: conclusion.tex
\section{Conclusion and Outlook}

In this paper, we propose a novel approach for scaling hand gestures such that CNNs can recognize without requiring an enormous quantity of training data or extra training effort. For this objective, we create and share a benchmark dataset, Scaled Hand Gestures Dataset (SHGD), which contains gesture tuples having a sequence of gesture phonemes. Moreover, we have proposed a network architecture for recognition of gesture tuples using a novel Viterbi-like decoder. In our experiments, we have used the 2D and 3D versions of the SqueezeNet and MobileNetV2 models. We achieve a classification accuracy of 98.47\% for 15 single gesture classes, and we achieve an accuracy of 94.69\% for recognition of 810 different \mbox{3-tuple} gesture classes.

The proposed approach contains utmost importance in order to meet the needs of applications requiring more complex HCI systems. We can easily scale hand gestures exponentially by increasing the number of gesture phonemes in multi-tuple gestures.  

Similar to Rotokas language (spoken on the island of Bougainville), which contains 11 phonemes, we plan to create a hand language by using multi-tuple gestures and start talking with our hands. 

%% file: acknowledgements.tex
\section*{Acknowledgements}
We gratefully acknowledge the support of NVIDIA Corporation with the donation of the Titan Xp GPU, and Infineon Technologies with the donation of Pico Monstar ToF camera used for this research.